\documentclass{article}
\usepackage{spconf,amsmath,graphicx}
\usepackage{amsmath}
\usepackage{amsfonts}
\usepackage{amsthm}
\usepackage{enumitem}
\newtheorem*{theorem}{Theorem}
\newtheorem*{definition}{Definition}
\usepackage{tikz}
\newcommand{\Arrow}{%
	\parbox{.1cm}{\tikz{\draw[->](0,0)--(.1cm,0);}}
}

\title{Does Your Model Think Like an Engineer? Explainable AI for Bearing Fault Detection with Deep Learning}

\name{Thomas Decker\textsuperscript{\rm 1,}\textsuperscript{\rm 2}, Michael Lebacher\textsuperscript{\rm 2}, Volker Tresp\textsuperscript{\rm 1,}\textsuperscript{\rm 2}}
\address{\textsuperscript{\rm 1}Ludwig Maximilians Universität,
Munich, Germany and \textsuperscript{\rm 2}Siemens AG, Munich, Germany}

\begin{document}
\ninept
\maketitle
\begin{abstract}
Deep Learning has already been successfully applied to analyze industrial sensor data in a variety of relevant use cases. However, the opaque nature of many well-performing methods poses a major obstacle for real-world deployment. Explainable AI (XAI) and especially feature attribution techniques promise to enable insights about how such models form their decision. But the plain application of such methods often fails to provide truly informative and problem-specific insights to domain experts. In this work, we focus on the specific task of detecting faults in rolling element bearings from vibration signals. We propose a novel and domain-specific feature attribution framework that allows us to evaluate how well the underlying logic of a model corresponds with expert reasoning. Utilizing the framework we are able to validate the trustworthiness and to successfully anticipate the generalization ability of different well-performing deep learning models. Our methodology demonstrates how signal processing tools can effectively be used to enhance Explainable AI techniques and acts as a template for similar problems. 
\end{abstract}

\section{Introduction}

\begin{figure}[t]
	\centering
	\includegraphics[width=0.9\columnwidth]{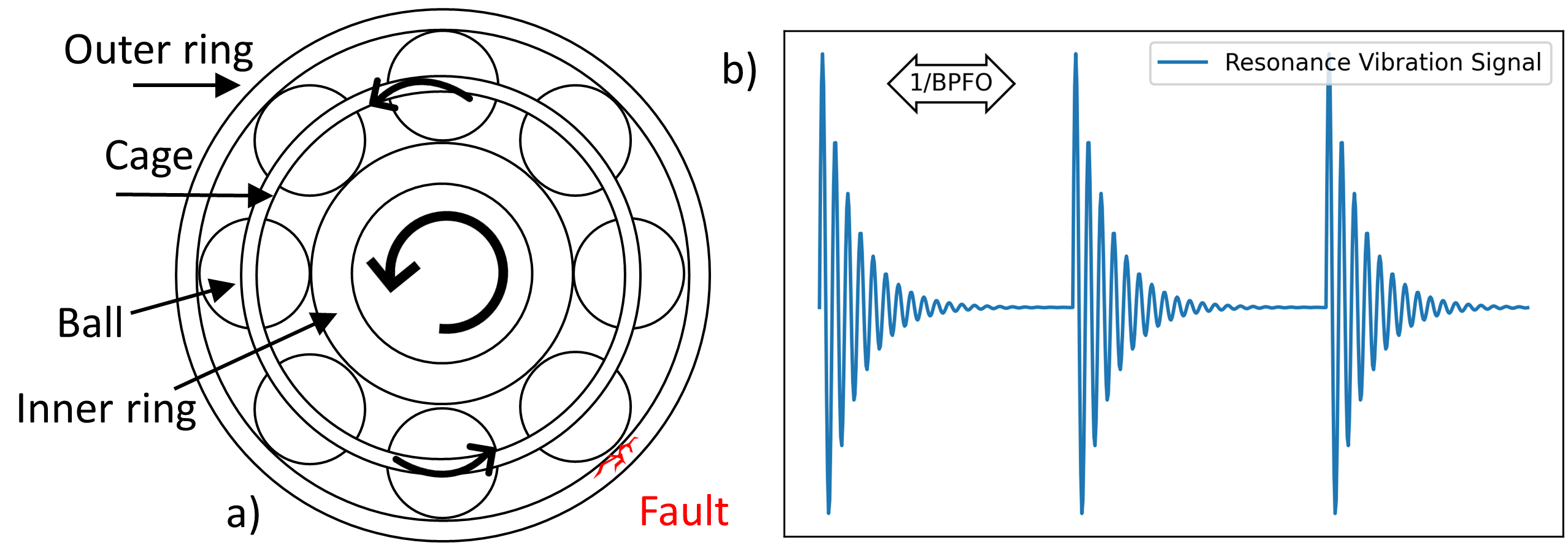}
	\caption{a) Illustration of a rolling element bearing with outer ring fault: During machine operation usually the outer ring is fixed, while the inner ring and the balls are rotating. This causes the balls to repeatedly hit the fault. b) A typical vibration signal induced by an outer ring fault where impacts occur with frequency BPFO.}
	\label{fig:1}
\end{figure}

Modern machine learning techniques such as deep learning have been applied to a variety of industry-relevant problems \cite{zhao2019deep}. A prominent example of such an industrial problem is the detection of faults in rolling element bearings (Figure \ref{fig:1}a), which are mechanical components that enable smooth movement with minimal friction and can be found in almost every rotating equipment. Estimates indicate that each year around 10 billion bearings are manufactured and roughly one billion are replaced for preventive reasons or due to failure \cite{SKF}. Thus, early detection of faulty bearings is a significant problem in industries and consequently, a large body of research in the past has been devoted to addressing it via elaborate signal modeling and processing \cite{randall2021vibration}. Although the logic of those approaches is well justified and rigorously grounded in the domain knowledge about the problem, their performance however is limited even in supervised experimental settings \cite{smith2015rolling}. On the other hand, various data-driven methods have demonstrated promising results in detecting bearing faults especially those based on deep learning \cite{jia2016deep}. Nevertheless, real-world deployment of such methods in industrial applications is generally impeded by the lack of transparency about how a model is forming its decision and whether it relies on the right reasons. This absence of explainability raises doubts regarding the actual validity and trustworthiness of the internal model logic as well as safety and legal concerns \cite{guidotti2018survey}. For this purpose, Explainable AI (XAI) \cite{doshi2017towards} and most prominently local feature attribution methods \cite{ras2018explanation} have been established as popular approaches to increase the transparency and trustworthiness of machine-based decisions. In essence, those methods usually try to quantify how important individual features are for a single specific output decision. While most of those methods can be motivated from a theoretical perspective, they often fail to provide truly informative insights for domain experts in practice \cite{bhatt2020explainable,alufaisan2021does}. In this work, we aim towards improving the quality and explanatory power of feature attribution methods when applied to industrial prediction problems focusing on the particular task of detecting faults in rolling element bearings from vibration signals. We propose a novel and problem-specific feature attribution framework and tailor it to the specific use case of bearing fault detection. It is based on the idea of appropriately translating the attributions into a feature space that is most informative for engineers and domain experts using appropriate signal processing techniques. This allows us to quantify post-hoc how well the logic of any trained model is aligned with classical reasoning. We also utilize the framework to conduct a comprehensive evaluation of different deep learning models, which all have successfully been trained to detect faulty bearings based on real vibration signals. Lastly, we demonstrate how the outcomes can be helpful to establish trust in the model's performance, aid model selection and to estimate the generalization ability to new machine operating conditions. 

\section{Background and related work}
\subsection{Modeling bearing fault vibrations} 
A bearing damage can be considered to excite characteristic resonance vibrations with a specific periodic signature depending on the nature of the fault. In this work, we will entirely focus on localized outer ring faults, but similar modeling can equivalently be done for other fault locations. Imagine the situation in Figure \ref{fig:1}a, where a bearing with a single local fault in the form of a crack at the outer ring is depicted. The classical way to model corresponding vibrations goes back to \cite{mcfadden1984model}. The bearing and attached machine parts can be considered a simple mechanical system and whenever a ball hits the defect during rotation, an impact impulse excites the system. An important observation is that the periodicity of impacts can physically be linked to the origin of the fault. For different locations, expected characteristic frequencies can be computed based on the rotation speed and the specific bearing geometry \cite{harris2001}. For outer ring faults this frequency is typically referred to as the Ball Passing Frequency Outer ring (BPFO), so if an outer ring fault is present, one can expect impact impulses to occur with the frequency BPFO. Each impulse will trigger a subsequent impulse response the typical structure of such fault-related vibration signals is also depicted in Figure \ref{fig:1}b. More details can be found in \cite{singh2015extensive}.
\begin{figure}[t]
	
	\centering
	\includegraphics[width=0.95\columnwidth]{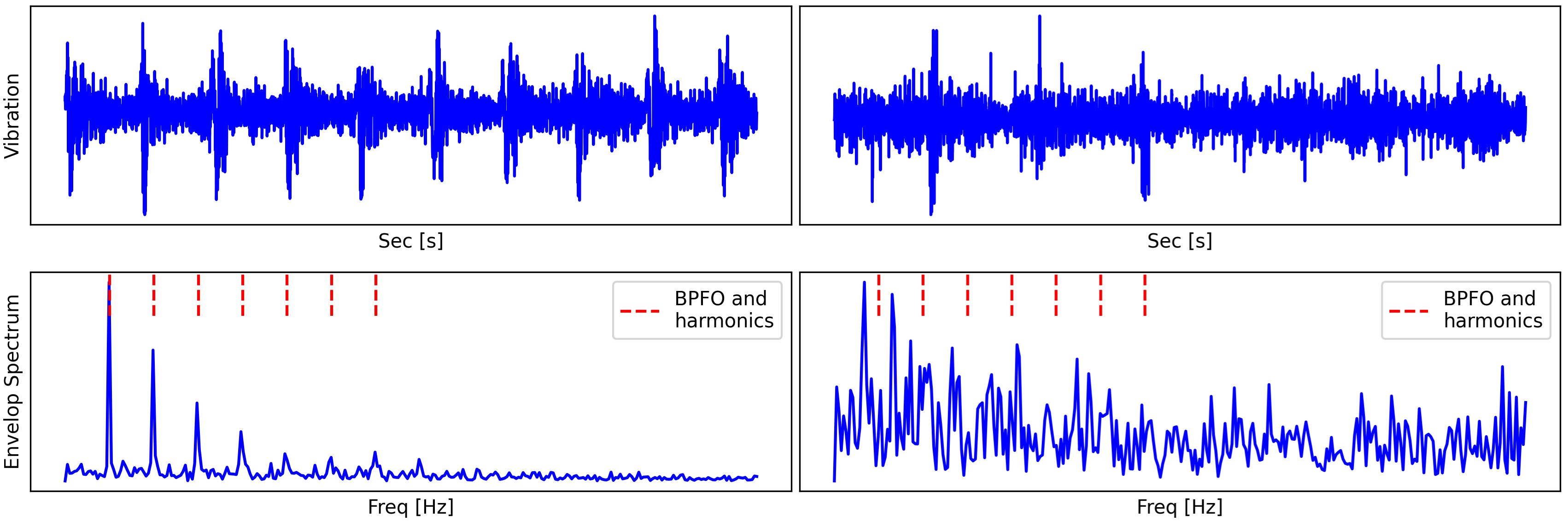}
	\caption{Two examples of envelope analysis: Top row shows two raw vibrations signals of a real outer ring fault with respective sections of the envelope spectra below. The fault on the left is detectable for domain experts via envelope analysis as strong peaks are present at the characteristic frequencies, while the right one is not.}
	\label{fig:3}
\end{figure}
\begin{figure}[t]
	\centering
	\includegraphics[width=0.99\columnwidth]{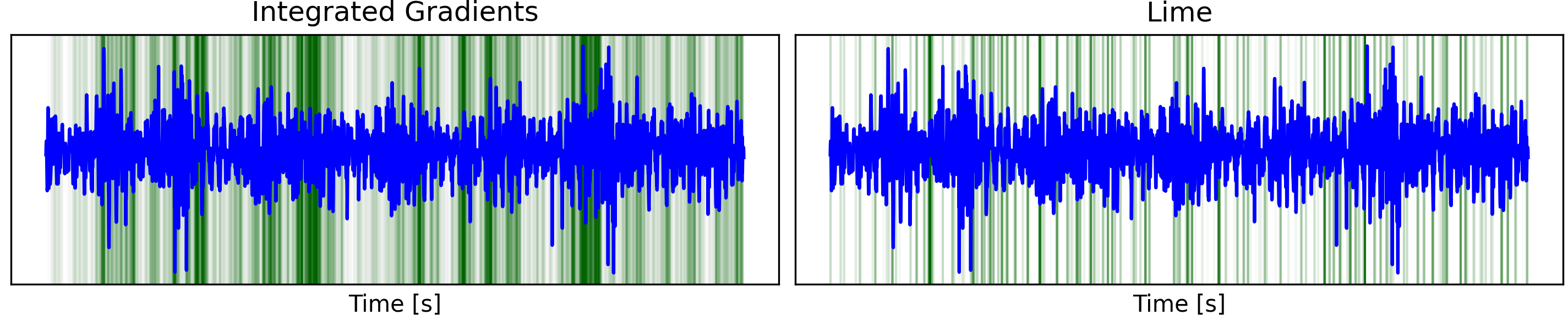}
	\caption{Results of two attribution methods in the time domain based on a convolutional neural network correctly identifying a fault.}
	\label{fig:4}
\end{figure}
\subsection{Detecting faults via envelope analysis and deep learning}
The application of envelope analysis to diagnose bearing faults has a long history and is still preferably used by domain experts in practice \cite{randall2011rolling, randall2021vibration}. If a specific defect is present in the bearing, it should reveal itself by producing a fault signal with a characteristic periodicity of impulses. To obtain such information it is advantageous to rather analyze a demodulated version of the signal such as the signal envelope. For this purpose, domain experts typically consider the analytic envelope which can be derived using the Hilbert transform $\mathcal{H}$ \cite{feldman2011hilbert}. After obtaining the envelope, diagnosing outer ring faults reduces to simply checking for significant peaks at the characteristic fault frequency BPFO and/or its harmonics in the corresponding envelope spectrum. In practice, envelope analysis is often combined with a variety of other signal processing techniques to enhance the presence of a potential fault signal \cite{randall2011rolling}. Nevertheless, evaluating raw signals often works sufficiently well and is considered as good baseline \cite{smith2015rolling}. While envelope analysis is rigorously motivated by physical domain knowledge about the problem, its performance might still be limited in practice, see Figure \ref{fig:3}. As a consequence, machine-learning-based approaches have been proposed as a solution by leveraging data \cite{liu2018artificial, jia2016deep}. In a recent survey  \cite{zhang2020deep}, 80 different deep learning approaches for bearing fault detection have been collected. The proposed models include a variety of different architectures and network types. They have mainly been trained on raw time or frequency representations of vibration signals and attain remarkably high performance. This raises the question of how such models are solving the task and if they are trustworthy enough for actual deployment. 

\subsection{Feature attribution methods}
Feature attribution methods try to quantify to which extent individual input features have contributed to a final output decision. In \cite{ancona2018towards}, a methodical distinction is made between perturbation and backpropagation-based approaches. Perturbation-based methods attribute importance to a feature by simulating its absence and aggregating the resulting predictions \cite{covert2021explaining}. Examples are LIME \cite{ribeiro2016should} or variations based on Shapley Values \cite{lundberg2017unified, sundararajan2020many}. Backpropagation-based methods typically require the model of interest to be differentiable as feature attributions are computed by aggregation or modification of gradients \cite{ancona2018towards}. To this category of methods belong, for instance, Integrated Gradients \cite{sundararajan2017axiomatic} or GradSHAP \cite{erion2021improving}. 
\section{Informative feature attribution for bearing fault detection}
\begin{figure}[t]
	\centering
	\includegraphics[width=1.\columnwidth]{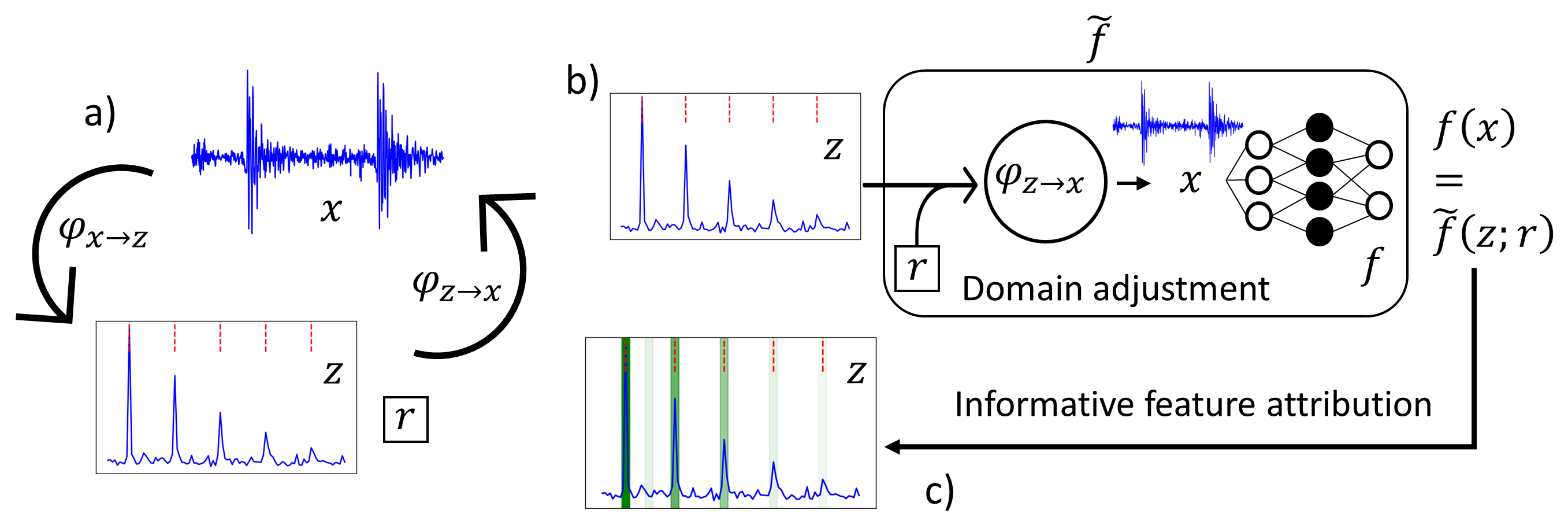}
	\caption{Using domain mappings (a) one can adjust a model $f$ trained on original features $x$ to work equivalently based on interpretable features $z$ yielding an augmented model $\tilde{f}(z;r)$ (b) which enables retrieval of informative attributions (c)}
	\label{fig:6}
\end{figure}

From a mathematical perspective the problem of local feature attribution can be stated in the following way: A decision of a machine learning model can be considered as a function $f: \mathbb{R}^d \rightarrow \mathbb{R}$ mapping input data $x \in \mathbb{R}^d$ to a real number $f(x)$ expressing a model prediction. The goal of feature attribution is to identify an importance vector $\phi \in \mathbb{R}^d$ such that $\phi_i$ quantifies the influence that each input feature $x_i$ had on the model prediction $f(x)$ for fixed input $x$. Typically any such method will retrieve importance scores in units of the raw input features $x$, which can significantly limit its explanatory power. Consider, for instance, a model trained on raw vibration recordings that classifies a signal as fault related. Figure \ref{fig:4}, displays typical attribution results in the original feature domain, where the color intensity indicates the importance of a signal value in time for the model prediction. Inferring a specific pattern or finding concrete reasons explaining how the model detected the fault is hardly possible. To mitigate this \cite{ribeiro2016should,singh2020transformation,de2020human} introduced the idea to rather compute feature importance with respect to an interpretable representation $z \in \mathcal{Z}$ of the explained instance $x \in \mathcal{X}$. In general, this requires a predefined mathematical setting to properly switch back and forth between the original and interpretable representations and we propose the following formulation:
\begin{definition}
	A domain mapping between an interpretable domain $\mathcal{Z}$ with residual component $R$ and a feature space $\mathcal{X}$ is specified by:
	\begin{align*}
		\varphi_{x\Arrow z}: \mathcal{X} \rightarrow \mathcal{Z} \times R \qquad \text{and} \qquad \varphi_{z\Arrow x}: \mathcal{Z} \times R \rightarrow \mathcal{X}
	\end{align*} such that for all $x$: $\quad \varphi_{z\Arrow x}(\varphi_{x\Arrow z}(x))=x$	
\end{definition} The purpose of the residual component $R$ is to restrict the evaluation of feature importance to certain parts of the representation, while still retaining sufficient information to recover the original input. To retrieve attributions based on the representation $z$ and residual $r$ one can simply consider the original model $f$ and adjust its input domain post-hoc. This yields an augmented model $\tilde{f}$ given by:
{\small
	\begin{align*}
		\tilde{f}(z;r)=f(\varphi_{z\Arrow x}(z,r))\quad \text{with} \quad (z, r) = \varphi_{x\Arrow z}(x)
\end{align*}}
This merely extends the original model to be specified in terms of a different input domain and keeps all characteristics of $f$ intact without any need for retraining. The idea is also illustrated in Figure \ref{fig:6}. 
In general, one seeks an interpretable representation that yields most informative feature attributions for a specific application. Thus, the natural choice when dealing with bearing vibrations is the envelope spectrum since it is the feature space that engineers analyze based on their domain knowledge. The implementation of such an attribution framework requires appropriate domain mappings translating between raw vibration signals and corresponding envelope spectra:
\begin{theorem}
	With Fourier Transform $\mathcal{F}$, a domain mapping between the time domain and the envelope spectrum is defined by:
	\begin{align*}
		&\varphi_{x\Arrow z}(x) = \Big(\mathcal{F}\big( \lvert x  + i\mathcal{H}(x) \rvert \big), \arg \big(x  + i\mathcal{H}(x) \big) \Big)  \\ &\varphi_{z\Arrow x}(z,r) = \mathcal{F}^{-1}(z)\cos(r)
	\end{align*}

\end{theorem}
Hence, to obtain the importance of individual components of the envelope spectrum $X_{\textit{envel}}$ one can simply apply feature attribution methods to an input domain adjusted model:
{\small
	\begin{align*}
		\tilde{f}^{\textit{envel}}(z;r)= f\left(\mathcal{F}^{-1}(X_{\textit{envel}})cos(r)\right)
\end{align*}}
Note that $\tilde{f}^{\textit{envel}}$ is only a function of the envelope spectrum as the residual $r$ is fixed and $\mathcal{F}^{-1}$ is linear. This means that the adjusted model only differs from the original one by an additional linear input transformation. In particular, if the considered model $f$ is a neural network this can be thought of as adding an extra linear layer without bias at the beginning. Hence, all feature attributions designed for deep neural networks are automatically applicable and well-defined for $\tilde{f}^{\textit{envel}}$ by treating the domain transformation as initial linear layer.
\section{Quantifying domain knowledge alignment}
By computing feature attributions based on the envelope spectrum as interpretable representation, we are able to evaluate which frequencies of the envelope spectrum are important for the model, although it was originally trained in a different domain. Since this is the same feature space that a domain expert would analyze manually, one can validate how well individual decisions of a deep learning model correspond with traditional domain expert reasoning. During diagnosis, domain experts focus their attention mainly to specific frequencies of the envelope spectrum and for outer ring faults, those are the BPFO and its multiples. Hence, we are treating the corresponding feature importance scores as informative content of an attribution result. Let $x$ be a signal for which we seek an explanation and  $\phi(x)$ be the corresponding result of a feature attribution method with respect to frequencies of the envelope spectrum. Now define $\mathcal{I}_m$ as index set such that $\phi_{\mathcal{I}_m}(x)$ captures the attribution allocated to BPFO and its $m$ harmonics. To quantify the alignment we propose two metrics. The Informative Attribution Share (IAS) simply captures how much of total attribution ends up in informative features:
\begin{itemize}[leftmargin=55pt]
	\item[ ] $\textit{IAS}(x) = \lVert\phi_{\mathcal{I}_m}(x)\rVert_1 / \lVert \phi(x) \rVert_1 $ 
\end{itemize}
 Although this quantity is intuitive, its magnitude could be undermined when $\lvert \mathcal{I}_m \rvert$ is very small compared to the total number of input features or $\phi$ is noisy. Therefore, we also propose as an alternative an Informative Signal to Noise Ratio (ISNR): 
\begin{itemize}
	\item[ ] $\textit{ISNR}(x) = 10 \log_{10}\left(\sum_{i \in \mathcal{I}_m} \phi_i^2(x) / \sum_{i \not\in \mathcal{I}_m} \phi_i^2(x)\right)$
\end{itemize}
By defining the $\phi_{\mathcal{I}_m}$ as informative signal content and the remaining attributions as uninformative noise we can leverage the concept of signal-to-noise ratios (in dB) to compute the relative energy contained in $\phi_{\mathcal{I}_m}$. Note that all these measures are defined instancewise and evaluate local feature attribution results for a specific model prediction of interest. To obtain overall metrics quantifying how well an entire model is aligned with domain knowledge we consider global versions $\mathbb{E}_{\mathcal{X}}[\textit{IAS}(x)]$ and $\mathbb{E}_{\mathcal{X}}[\textit{ISNR}(x)]$ as expectation over the data distribution, which can be approximated via samples. 
\begin{table}[t]
	\small
	\centering
	\begin{tabular}{|c|cc|cc|cc|}
		\hline
		& \multicolumn{2}{c|}{Train} &  \multicolumn{2}{c|}{Real}  & \multicolumn{2}{c|}{Generalize} \\ 
		\hline
		Model & Prec  & Rec  & Prec& Rec  & Prec  & Rec  \\  
		\hline
		CNN & 0.992  & 0.999& 0.988& 0.956  & 0.994 & 0.581 \\ 
		FCN & 0.980  & 0.996 & 0.958& 0.639&  0.487 & 0.462 \\
		LSTM & 0.995  & 0.993& 0.991& 0.814& 0.996 & 0.524 \\
		BASE &0.949 &0.815& 0.701 &0.751 &0.657 & 0.844 \\ 
		\hline
	\end{tabular}
	\caption{Precision (Prec) and recall (Rec) of all fault detection models on different Paderborn datasets.}  
	\label{tab:2}
\end{table}
\section{Numerical experiments}
\subsection{Dataset and model fitting}
To demonstrate the utility of the introduced alignment metrics we conduct numerical experiments based on real vibration signals. The Kat-DataCenter of Paderborn University has published an extensive dataset containing vibration recordings of healthy and outer ring damaged bearings \cite{lessmeier2016condition}. The data results from experiments with 6 undamaged bearing, 6 bearings with artificially damaged outer rings and 5 bearings with genuine outer ring faults created via accelerated lifetime tests. For each bearing, also variable operating conditions including two different rotation speeds have been taken into account. This makes the dataset well-suited to evaluate the performance and generalization capabilities of fault detection models under authentic circumstances. All provided signals are recorded with a sampling rate of 64kHz and we split all recordings into individual signals of length $d=16000$ normalized to be between $-1$ and $1$. We further trained three different neural network architectures to distinguish between healthy and outer ring fault-related vibrations. The considered models consist of either a Convolutional Neural Network (CNN), a Fully-Connected Network (FCN) or a Long Short-Term Memory Network (LSTM) as main feature extractor followed by subsequent dense classification layers. We also created a baseline classifier (BASE) as a crude approximation of how domain experts would perform based on standard envelope analysis. This classifier computes for each signal its envelope spectrum and identifies a fault if the share of energy contained in the characteristic frequency BPFO and its harmonics exceeds a certain threshold. The optimal threshold was determined individually for each dataset by maximizing the F1 score via grid search. To evaluate all models, we consider distinctive train and test scenarios replicating two authentic real-world generalization challenges. We train all models to classify healthy vibrations and those of artificially induced faults obtained under fixed rotation speed (rpm = $1500$). This training dataset contains $10950$ signals and about 52\% faults. We first test the fault detection performance of these models when distinguishing $4059$ realistic fault signals from undamaged vibrations. We further evaluate how the models would generalize to a dataset comprised of healthy and realistic fault signals arising from unseen rotation speed conditions (rpm = $900$). This generalization set contains $3070$ signals with 45\% faults. The resulting performance metrics are summarized in Table \ref{tab:2}. The simple BASE model achieves reasonably good results indicating that the envelope spectrum is indeed a favorable feature space with discriminative power but is outperformed by all neural networks on the training data. LSTM and especially CNN maintain high detection accuracy when facing realistic signals surpassing BASE, while FCN only detects $64\%$ of the faults. Under unseen operating conditions, all deep learning models face a decline in performance. Interestingly, CNN and LSTM still exhibit high precision but fail to detect all available faults, whereas FCN shows virtually no predictive power.
\subsection{Establishing trust and anticipating generalization ability}
\begin{figure}[t]
	\centering
	\includegraphics[width=0.99\columnwidth]{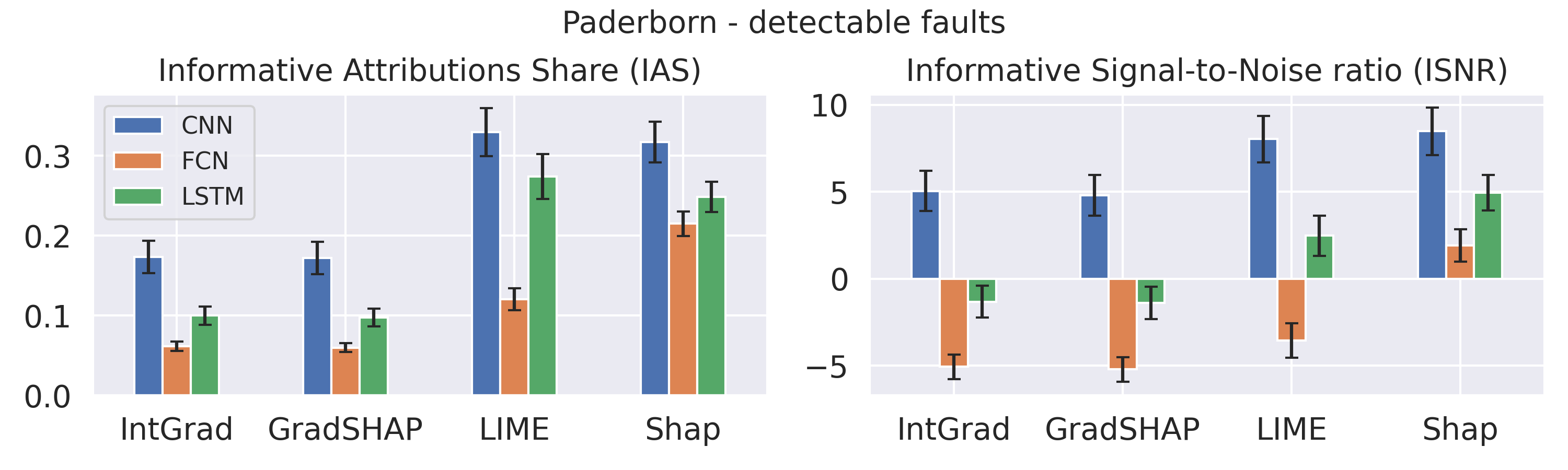}
	\includegraphics[width=0.99\columnwidth]{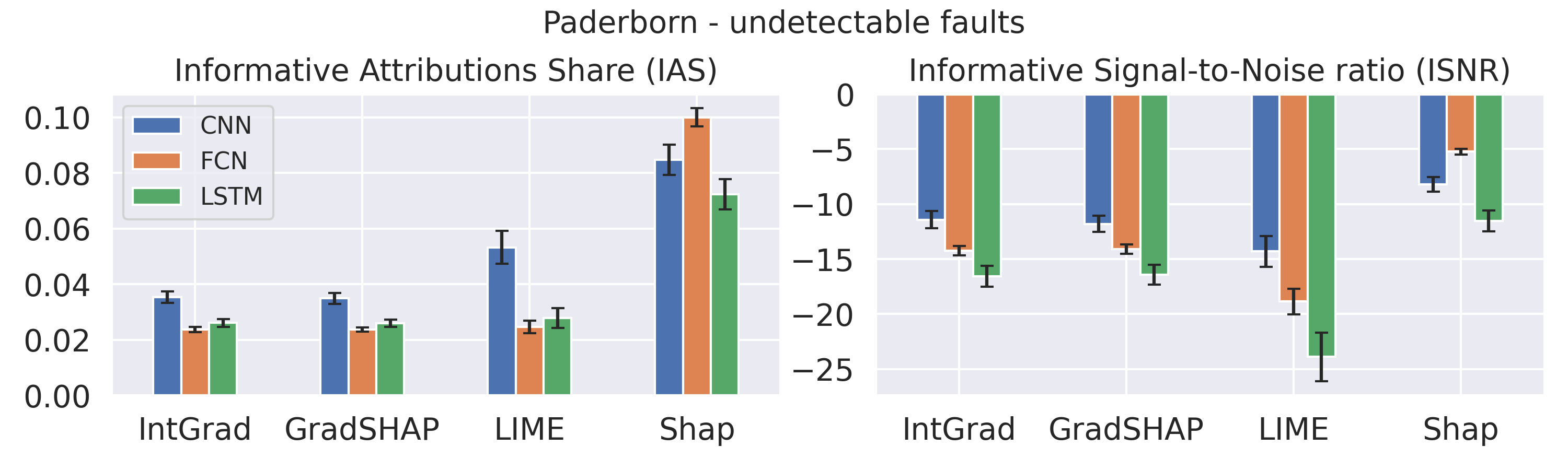}
	\includegraphics[width=0.99\columnwidth]{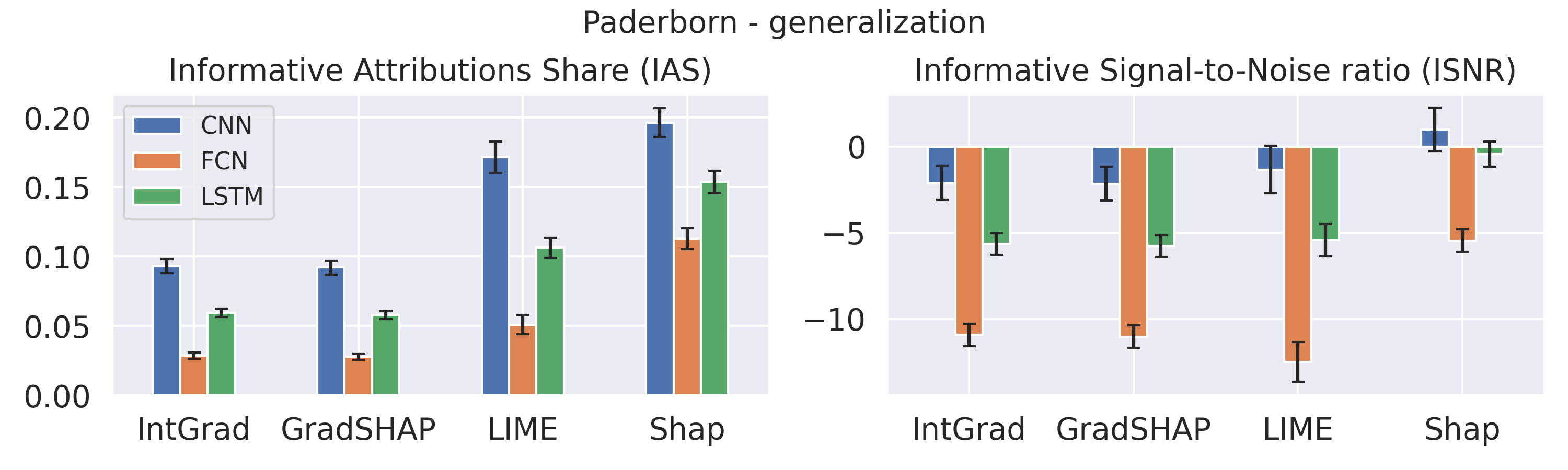}
	\includegraphics[width=0.93\columnwidth]{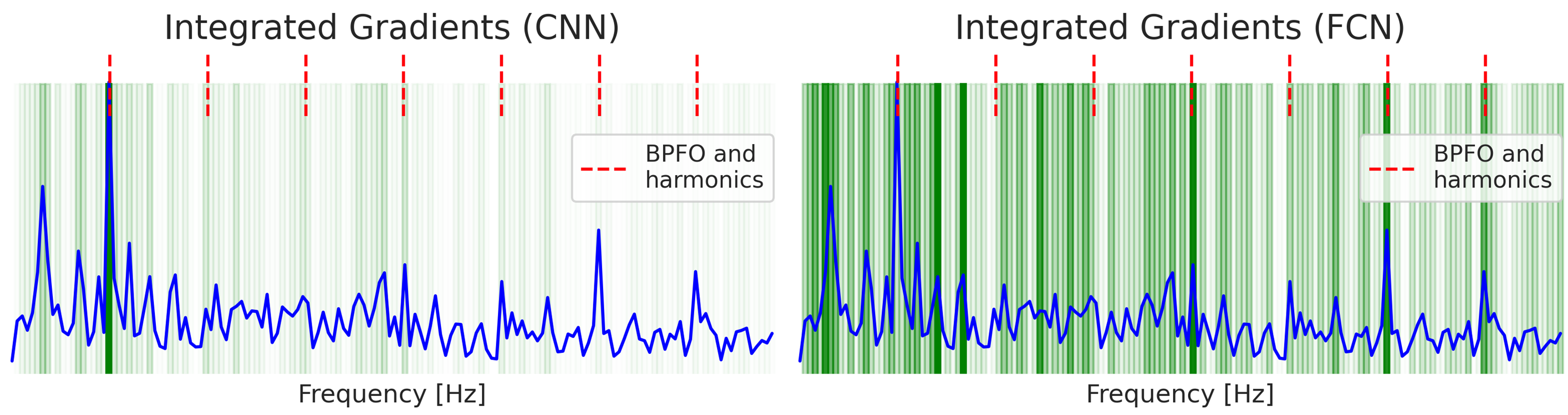 }
	\caption{Global domain knowledge alignment results on detectable (top row) and undetectable faults (second row) of the training data and on predicted faults from the generalization set (third row) including 95\% confidence levels as error bars. Bottom row: relevant section of local attribution results in the envelope spectrum of a realistic fault signal (same as in Figure \ref{fig:4}) which is correctly identified by the CNN focusing on the BPFO but misclassified by the FCN.}
	\label{fig:5}
\end{figure}

From an engineer's perspective, a trustworthy bearing fault detection model shall detect all faults that show classical symptoms in the envelope spectrum based on these characteristics and still be able to perform well on faults that are undetectable to experts. Thus, when choosing the particular model to be deployed among similar well-performing ones, the one with higher domain knowledge alignment on detectable faults is preferable. To quantify domain knowledge alignment we define the BPFO and its 10 harmonics as informative features. When dealing with real signals in practice, such values can be expected to deviate by up to 2\% \cite{randall2011rolling}, so we consider all frequency features within that range. To retrieve global metrics we used $200$ signals from the training set as well as $200$ signals from the generalization set that have been classified as a fault separately for each model. To retrieve the underlying informative feature attributions with respect to the envelope spectrum we utilize in total four different methods to ensure conclusive results. We choose Integrated Gradients (IntGrad) and GradSHAP as backpropagation-based methods, while Shapley Values (Shap) and LIME serve as popular perturbation-based methods. The computation of all methods has been conducted via Captum \cite{kokhlikyan2020captum}. The resulting outcomes are summarized in Figure \ref{fig:5} and enable relevant insights regarding the models' trustworthiness and generalization ability.
 For trust-based model selection on the Paderborn dataset we compute global domain knowledge alignment separately for detectable and non-detectable faults (Figure \ref{fig:5}). This distinction is based on the baseline classifier. Comparing the individual metrics immediately reveals that all models have substantially higher scores on detectable faults. In particular, the low IAS and negative ISNR levels for undetectable faults imply that models do indeed utilize different characteristics to detect faults without classic symptoms. To pick the most trustworthy model one can consult the alignment results on detectable faults, where again the CNN is significantly superior, followed by the LSTM. Moreover, the order of domain knowledge correspondence on detectable faults also matches exactly the associated performances on realistic faults in Table \ref{tab:2}. Hence, models that are more in line with expert reasoning on detectable faults also generalize better when switching from artificial conditions to authentic faults. This can also be verified for individual predictions. In Figure \ref{fig:5} (bottom row), local informative attribution results of the same signal considered in Figure \ref{fig:4} are presented. While raw attributions in the time domain are incomprehensible, the interpretable versions reveal that the CNN identifies the fault mainly based on the spike present at the BPFO frequency like an expert would. The FCN however fails to detect the fault while focusing on several peaks unrelated to the characteristic frequencies. Also, the robustness with respect to new machine operating conditions can be estimated via our framework. In Figure \ref{fig:5} (third row), the alignment metrics indicate to which extent models rely on domain knowledge when predicting a fault in the generalization set. While the overall level of the metrics is rather low, CNN and LSTM seem to rely significantly more on domain expertise when predicting a fault compared to FCN. This once again coincides with the actual performances in Table \ref{tab:2} where all reported metrics deteriorated but CNN and LSTM maintain high precision. 

\section{CONCLUSION}
In this work, we proposed a framework that makes the opaque nature of machine learning models more accessible to domain experts in a novel way and enables valuable insights regarding the trustworthiness and generalization ability. Although we primarily dealt with the concrete implementation for the specific use case of bearing fault detection, the methodology is equivalently applicable to other problems as long as an informative feature space with corresponding domain maps exist. Despite the lack of general principles regarding what really makes an explanation informative \cite{lage2019evaluation}, we believe that use case specific attributions and quantifying to which extent a model thinks like an engineer constitutes a step in the right direction.

\bibliographystyle{IEEEbib}
\bibliography{icassp.bib}

\begin{thebibliography}{10}

\bibitem{zhao2019deep}
Rui Zhao, Ruqiang Yan, Zhenghua Chen, Kezhi Mao, Peng Wang, and Robert~X Gao,
\newblock ``Deep learning and its applications to machine health monitoring,''
\newblock {\em Mechanical Systems and Signal Processing}, vol. 115, pp.
  213--237, 2019.

\bibitem{SKF}
SKFGroup,
\newblock ``Bearing damage and failure analysis,'' 2017,
\newblock Accessed: 2022-04-06.

\bibitem{randall2021vibration}
Robert~Bond Randall,
\newblock {\em Vibration-based condition monitoring: industrial, automotive and
  aerospace applications},
\newblock John Wiley \& Sons, 2021.

\bibitem{smith2015rolling}
Wade~A Smith and Robert~B Randall,
\newblock ``Rolling element bearing diagnostics using the case western reserve
  university data: A benchmark study,''
\newblock {\em Mechanical systems and signal processing}, vol. 64, pp.
  100--131, 2015.

\bibitem{jia2016deep}
Feng Jia, Yaguo Lei, Jing Lin, Xin Zhou, and Na~Lu,
\newblock ``Deep neural networks: A promising tool for fault characteristic
  mining and intelligent diagnosis of rotating machinery with massive data,''
\newblock {\em Mechanical systems and signal processing}, vol. 72, pp.
  303--315, 2016.

\bibitem{guidotti2018survey}
Riccardo Guidotti, Anna Monreale, Salvatore Ruggieri, Franco Turini, Fosca
  Giannotti, and Dino Pedreschi,
\newblock ``A survey of methods for explaining black box models,''
\newblock {\em ACM computing surveys (CSUR)}, vol. 51, no. 5, pp. 1--42, 2018.

\bibitem{doshi2017towards}
Finale Doshi-Velez and Been Kim,
\newblock ``Towards a rigorous science of interpretable machine learning,''
\newblock {\em arXiv preprint arXiv:1702.08608}, 2017.

\bibitem{ras2018explanation}
Gabri{\"e}lle Ras, Marcel van Gerven, and Pim Haselager,
\newblock ``Explanation methods in deep learning: Users, values, concerns and
  challenges,''
\newblock in {\em Explainable and interpretable models in computer vision and
  machine learning}, pp. 19--36. Springer, 2018.

\bibitem{bhatt2020explainable}
Umang Bhatt, Alice Xiang, Shubham Sharma, Adrian Weller, Ankur Taly, Yunhan
  Jia, Joydeep Ghosh, Ruchir Puri, Jos{\'e}~MF Moura, and Peter Eckersley,
\newblock ``Explainable machine learning in deployment,''
\newblock in {\em Proceedings of the 2020 conference on fairness,
  accountability, and transparency}, 2020, pp. 648--657.

\bibitem{alufaisan2021does}
Yasmeen Alufaisan, Laura~R Marusich, Jonathan~Z Bakdash, Yan Zhou, and Murat
  Kantarcioglu,
\newblock ``Does explainable artificial intelligence improve human
  decision-making?,''
\newblock in {\em Proceedings of the AAAI Conference on Artificial
  Intelligence}, 2021, vol.~35, pp. 6618--6626.

\bibitem{mcfadden1984model}
PD~McFadden and JD~Smith,
\newblock ``Model for the vibration produced by a single point defect in a
  rolling element bearing,''
\newblock {\em Journal of sound and vibration}, vol. 96, no. 1, pp. 69--82,
  1984.

\bibitem{harris2001}
Tedric~A. Harris,
\newblock {\em Rolling Bearing Analysis},
\newblock John Wiley \& Sons, New York, 2001.

\bibitem{singh2015extensive}
Sarabjeet Singh, Carl~Q Howard, and Colin~H Hansen,
\newblock ``An extensive review of vibration modelling of rolling element
  bearings with localised and extended defects,''
\newblock {\em Journal of Sound and Vibration}, vol. 357, pp. 300--330, 2015.

\bibitem{randall2011rolling}
Robert~B Randall and Jerome Antoni,
\newblock ``Rolling element bearing diagnostics—a tutorial,''
\newblock {\em Mechanical systems and signal processing}, vol. 25, no. 2, pp.
  485--520, 2011.

\bibitem{feldman2011hilbert}
Michael Feldman,
\newblock ``Hilbert transform in vibration analysis,''
\newblock {\em Mechanical systems and signal processing}, vol. 25, no. 3, pp.
  735--802, 2011.

\bibitem{liu2018artificial}
Ruonan Liu, Boyuan Yang, Enrico Zio, and Xuefeng Chen,
\newblock ``Artificial intelligence for fault diagnosis of rotating machinery:
  A review,''
\newblock {\em Mechanical Systems and Signal Processing}, vol. 108, pp. 33--47,
  2018.

\bibitem{zhang2020deep}
Shen Zhang, Shibo Zhang, Bingnan Wang, and Thomas~G Habetler,
\newblock ``Deep learning algorithms for bearing fault diagnostics—a
  comprehensive review,''
\newblock {\em IEEE Access}, vol. 8, pp. 29857--29881, 2020.

\bibitem{ancona2018towards}
Marco Ancona, Enea Ceolini, Cengiz {\"O}ztireli, and Markus Gross,
\newblock ``Towards better understanding of gradient-based attribution methods
  for deep neural networks,''
\newblock in {\em 6th International Conference on Learning Representations
  (ICLR)}, 2018.

\bibitem{covert2021explaining}
Ian Covert, Scott Lundberg, and Su-In Lee,
\newblock ``Explaining by removing: A unified framework for model
  explanation,''
\newblock {\em Journal of Machine Learning Research}, vol. 22, no. 209, pp.
  1--90, 2021.

\bibitem{ribeiro2016should}
Marco~Tulio Ribeiro, Sameer Singh, and Carlos Guestrin,
\newblock ``" why should i trust you?" explaining the predictions of any
  classifier,''
\newblock in {\em Proceedings of the 22nd ACM SIGKDD international conference
  on knowledge discovery and data mining}, 2016, pp. 1135--1144.

\bibitem{lundberg2017unified}
Scott~M Lundberg and Su-In Lee,
\newblock ``A unified approach to interpreting model predictions,''
\newblock {\em Advances in neural information processing systems}, vol. 30,
  2017.

\bibitem{sundararajan2020many}
Mukund Sundararajan and Amir Najmi,
\newblock ``The many shapley values for model explanation,''
\newblock in {\em International conference on machine learning}. PMLR, 2020,
  pp. 9269--9278.

\bibitem{sundararajan2017axiomatic}
Mukund Sundararajan, Ankur Taly, and Qiqi Yan,
\newblock ``Axiomatic attribution for deep networks,''
\newblock in {\em International conference on machine learning}. PMLR, 2017,
  pp. 3319--3328.

\bibitem{erion2021improving}
Gabriel Erion, Joseph~D Janizek, Pascal Sturmfels, Scott~M Lundberg, and Su-In
  Lee,
\newblock ``Improving performance of deep learning models with axiomatic
  attribution priors and expected gradients,''
\newblock {\em Nature machine intelligence}, vol. 3, no. 7, pp. 620--631, 2021.

\bibitem{singh2020transformation}
Chandan Singh, Wooseok Ha, Francois Lanusse, Vanessa Boehm, Jia Liu, and Bin
  Yu,
\newblock ``Transformation importance with applications to cosmology,''
\newblock {\em arXiv preprint arXiv:2003.01926}, 2020.

\bibitem{de2020human}
Damien de~Mijolla, Christopher Frye, Markus Kunesch, John Mansir, and Ilya
  Feige,
\newblock ``Human-interpretable model explainability on high-dimensional
  data,''
\newblock {\em arXiv preprint arXiv:2010.07384}, 2020.

\bibitem{lessmeier2016condition}
Christian Lessmeier, James~Kuria Kimotho, Detmar Zimmer, and Walter Sextro,
\newblock ``Condition monitoring of bearing damage in electromechanical drive
  systems by using motor current signals of electric motors: A benchmark data
  set for data-driven classification,''
\newblock in {\em PHM Society European Conference}, 2016, vol.~3.

\bibitem{kokhlikyan2020captum}
Narine Kokhlikyan, Vivek Miglani, Miguel Martin, Edward Wang, Bilal Alsallakh,
  Jonathan Reynolds, Alexander Melnikov, Natalia Kliushkina, Carlos Araya, Siqi
  Yan, et~al.,
\newblock ``Captum: A unified and generic model interpretability library for
  pytorch,''
\newblock {\em arXiv preprint arXiv:2009.07896}, 2020.

\bibitem{lage2019evaluation}
Isaac Lage, Emily Chen, Jeffrey He, Menaka Narayanan, Been Kim, Sam Gershman,
  and Finale Doshi-Velez,
\newblock ``An evaluation of the human-interpretability of explanation,''
\newblock {\em arXiv preprint arXiv:1902.00006}, 2019.

\end{thebibliography}

\end{document}